\documentclass[10pt,twocolumn,letterpaper]{article}

\usepackage[pagenumbers]{cvpr} 

\usepackage{graphicx}
\usepackage{amsmath}
\usepackage{amssymb}
\usepackage{booktabs}
\usepackage{multirow}
\usepackage{enumitem}

\usepackage[pagebackref,breaklinks,colorlinks]{hyperref}

\usepackage[capitalize]{cleveref}
\crefname{section}{Sec.}{Secs.}
\Crefname{section}{Section}{Sections}
\Crefname{table}{Table}{Tables}
\crefname{table}{Tab.}{Tabs.}

\def\cvprPaperID{*****} 
\def\confName{CVPR}
\def\confYear{2023}

\begin{document}

\title{Adversarially-Guided Portrait Matting
}

\author{Sergej Chicherin\\
SaluteDevices\\
Moscow, Russia\\
{\tt\small serenkiy@ieee.org}
\and
Karen Efremyan\\
SaluteDevices, MIPT\\
Moscow, Russia\\
{\tt\small efremyan.kk@phystech.edu}
}
\maketitle
\begin{abstract}

We present a method for generating alpha mattes using a limited data source. We pretrain a novel transformer-based model (StyleMatte) on portrait datasets. We utilize this model to provide image-mask pairs for the StyleGAN3-based network (StyleMatteGAN). This network is trained unsupervisedly and generates previously unseen image-mask training pairs that are fed back to StyleMatte. We demonstrate that the performance of the matte pulling network improves during this cycle and obtains top results on the used human portraits and state-of-the-art metrics on animals dataset. Furthermore, StyleMatteGAN provides high-resolution, privacy-preserving portraits with alpha mattes, making it suitable for various image composition tasks. Our code is available at \href{https://github.com/chroneus/stylematte}{https://github.com/chroneus/stylematte}.
\end{abstract}
\section{Introduction}
\label{sec:intro}
Image matting is a classical computer vision problem~\cite{smith1996blue} of foreground object extraction by providing an additional $\alpha$-mask,  where $\alpha$ is the object intensity.
In other words, considering the composite image  as $C \in \mathbb{R} ^{3 \times  H \times W}$, foreground as $F$, background as $B$ and alpha matte value as $\alpha$ we have the following alpha-blending equation:
\begin{equation} \label{eq:alpha}
    C_{ij} = \alpha_{ij}*F_{ij}+(1-\alpha_{ij})*B_{ij} \quad \{i,j\} \in \{W, H\}
\end{equation}
When alpha equals 0 or 1, the composite turns into either the background or foreground, and the solution is the same as that for segmentation. 

The problem of pulling a matte using a given composing image is ill-posed in this interpretation. Given $C$ and estimation $B$, the equation~\eqref{eq:alpha} leads to an infinite solutions.
The matting was introduced in the early movie industry, where the special process of shooting scenes with fixed color backgrounds was utilized for decades. Usually, the ``chromakey'' or green screen pavilion is used as a background. In early studies, the solutions were based on various assumptions regarding the $F$, $B$, or $\alpha$ distribution.
Another tool for improving matting is trimap, a handwritten segmentation mask containing foreground, background, and unknown grey-marked areas.
In terms of \eqref{eq:alpha}, trimap approximate $\alpha$ mask with exact boundary conditions of 0 and 1 and a grey mask in $(0, 1)$. Various techniques have been proposed to propagate distributions from known background and foreground areas to unknown areas near mask boundaries.
\begin{figure}
  \centering
\includegraphics[width=\linewidth]{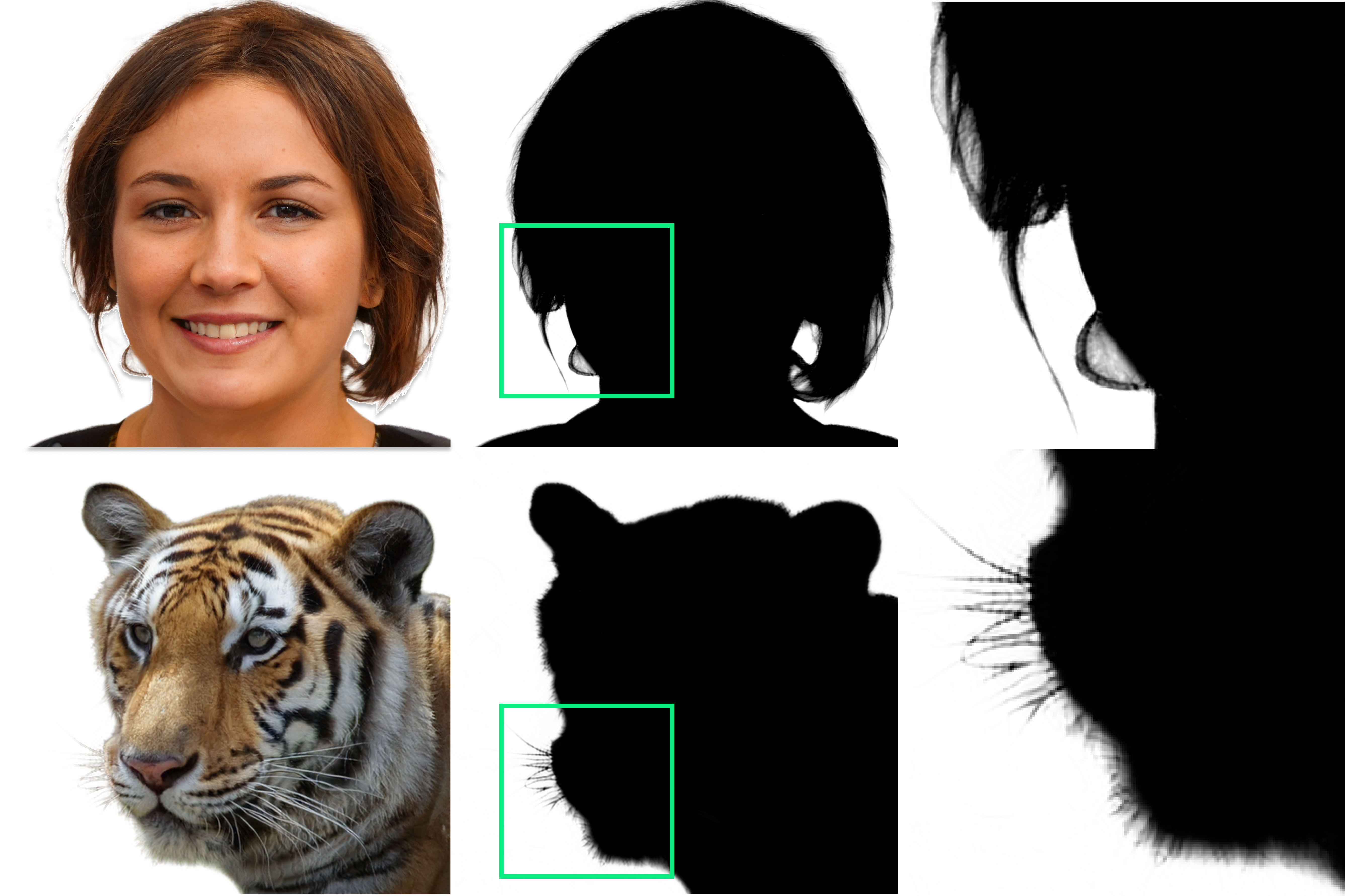}
  \caption{Example pairs of images and mattes produced by StyleMatteGAN. From left to right: image, alpha matte and its crop.}
  \label{fig: example}
\end{figure}
The difficulty of portrait matting comes from two different cases. The first one is connected with transparent objects and objects with holes like glasses(``true matting''). The second is related to image discretization. We can be certain about the binary border in high-resolution images that separates the foreground from the background. However, due to the discrete approximation, the border pixels are interpolated using information from both sides. In other words, the ``border matting'' is the same as segmentation at higher resolution and interpolation. This is particularly important for the hair pixels. If we take the upper bound of hair thickness as 1e-4 m and consider face image resolution in FFHQ dataset as 1024 px // 0.3 m  $\simeq$ 3000 px/m, then the rough estimation shows that each hair could be up to 0.3 px thick in HD photo. These subpixels are merged with the background, which leads to the fusion of the background and foreground.

Recent progress in neural networks has helped to automate matte pulling. The coarse-grain segmentation allows building trimap-free solutions in \cite{alphamatte,deepportrait}. Visual transformers~\cite{dosovitskiy2020image} significantly improve the semantic segmentation results ~\cite{segformer,dosovitskiy2020image}. However, this cannot be easily extrapolated to matting because of several difficulties. One of the difficulties is the limited availability of well-prepared images with alpha mattes.
At the time of writing there were only two datasets for training in open access - AM-2k~\cite{li2022matting} and P3M-10k
~\cite{vitae}. AM-2k contains 2000 high-resolution natural animal images from 20 categories with alpha mattes. P3M-10k contains 10421 privacy-preserving portraits with manually labeled mattes. In visual study, we find that the matte estimation in both datasets is noisy.
It is hard to obtain the ground-truth matting mask as a solution to equation \eqref{eq:alpha}. There are also other private datasets collected in a studio using a green screen setup and have license restrictions. Several works ~\cite{vitae,sengupta2020background} containing portrait images could not be shared due to privacy preservation. In the proposed work, we introduce a new matting network StyleMatteGAN which generates synthetic portraits with masks (Figure ~\ref{fig: example}).

To generate synthetic data, we take advantage of alias-free generative adversarial network~\cite{stylegan3}, which is referred to as StyleGAN3. StyleGAN3~\cite{stylegan3} avoids texture sticking on hairs, whiskers, and fur compared with its predecessors.
In brief, our contribution is:
  
\begin{enumerate}[topsep=0pt,itemsep=0pt,partopsep=0pt, parsep=0pt]

\item We designed a simple modern transformer-based model to pull the matte from the portrait image. This network achieves top results on the P3M-10k and AM-2k datasets.\item We studied modifications of StyleGAN3 to train on portraits with alpha masks unsupervisedly. These masks are consistent with the portraits generated by a single network. The distribution of the generated images is close to the original, with FID score below 6.\item We propose a method to improve the aforementioned networks using each other's outputs as new training samples.
\end{enumerate}

\section{Related Work}

\subsection{Image Matting}

It is customary to divide image matting approaches into two significant sections: trimap-based and trimap-free matting.

\subsection{Trimap-based matting methods}
As far as equation~\eqref{eq:alpha} is ill-posed, early works use additional inputs called trimaps. It is an auxiliary input that contains three regions: the pure foreground, pure background, and unknown area. The main idea is to estimate the foreground probability inside an unknown area using only the pure background ($\alpha = 0)$ and foreground ($\alpha = 1$) information.

One modern solution is MatteFormer~\cite{Park_2022_CVPR}, which represents the idea of using a trimap in transformers. Specifically, the authors generate trimap-based prior tokens, concatenate them to the local spatial tokens and use them in the self-attention mechanism. 

Another approach was proposed by \cite{mgm} using a guidance mask, which is a coarse-grain binary approximation to segmentation.
However, trimap generation is an additional human-guided task. Therefore, many practical solutions attempt to solve the matting task with only a given composite image as input.

We also found that the matting results strongly depend on the unknown area shape in the trimaps. We attempted different placements of unknown areas in the trimaps, and it was revealed that outputs are very sensitive to minor changes.

\begin{figure*}
  \centering
  \includegraphics[scale=0.5]{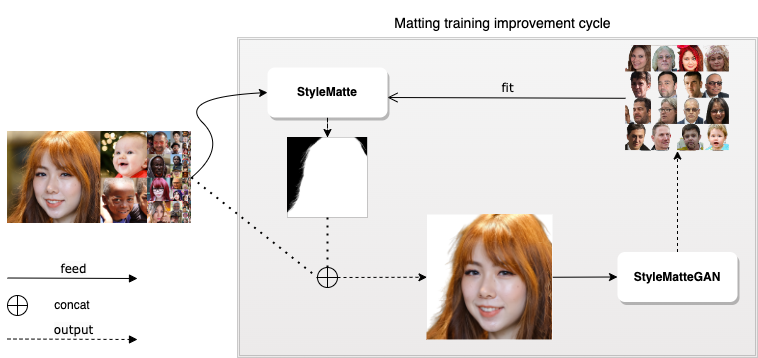}
  \caption{The highlighted cycle illustrates the matting improvement process.  Each iteration utilizes the current StyleMatte results to train StyleMatteGAN for synthetic matting composites generation. The synthetic dataset has been used for StyleMatte training.}
  \label{fig: improve}
\end{figure*}

\subsection{Trimap-free matting methods}

It is much more challenging to pull the matte without auxiliary guidance input. To solve equation~\eqref{eq:alpha}, we impose additional constraints. Some studies add constraints on the type of foreground objects, for example, portraits~\cite{rvm, MODNet, p3m}. In general, simply feeding RGB images to a single net may lead to artifacts in the predicted masks because it does not consider the semantic gap between the foreground and background, due to a lack of additional information.

In ``The World is Your Green Screen''~\cite{sengupta2020background}, the background image is pre-captured and used instead of a solid background color assumption.

In subsequent work, Robust Video Matting~\cite{rvm}, there is no need for background image capturing. Unlike most existing methods that perform video matting frame-by-frame as in-dependent images, RVM uses a recurrent architecture
to exploit the temporal information in videos.

MODNet~\cite{MODNet} presents a lightweight matting objective decomposition network for portrait matting in real-time with an input image. It utilizes an Efficient Atrous Spatial Pyramid Pooling (e-ASPP) module to fuse multi-scale features for semantic estimation, and a self-supervised sub-objective consistency strategy to adapt the network to address the domain shift problem.

Another novel approach is presented in ~\cite{alphamatte} which uses separate encoders to pull image context information and refine segmentation borders with common decoder. As an alternative to trimap-guided inputs, the authors use portrait segmentation from DeepLabV3 to feed the segmentation map to the encoder. Despite the state-of-the-art metrics that were achieved, this model is sensitive to the generated segmentation maps.

The End-to-end Deep Image Matting \cite{li2022matting} presents similar approach. They use a shared encoder ahead of the shared decoder and two parallel decoders for the local alpha matte and glance segmentation map estimation with the final aggregation head. They also published high-quality animal matting and background datasets alongside their networks. The shared encoder architecture was also utilized by Privacy-Preserving Portrait Matting \cite{vitae} which represents P3M-10k dataset that consists of 10000 high-resolution face-blurred portrait images with high-quality alpha mattes. We used aforementioned manually annotated AM-2k, P3M-10k and BG-20k background datasets for matting retrieval and background replacement experiments.

\subsection{Mask generation adversarial networks}
A small number of studies have attempted to generate synthetic matting masks. This idea originated from networks that generate segmentation and other masks ~\cite{3d-aware}. \cite{semanticGAN} uses GAN for semantic segmentation, building their model on StyleGAN2 in a semi-supervised manner. This approach allows for out-of-domain generalization, such as transferring from CT to MRI in medical imaging and real faces to paintings, sculptures, cartoons and animal faces.

AlphaGAN~\cite{lutz2018alphagan} proposes an adversarial network trained to predict alpha masks in conformity with adversarial loss that was trained to classify well-composited images. It comprises one generator $G$ and one discriminator $D$. G inputs an image composited from the foreground, alpha, and a random background concatenated with the trimap as 4-th channel in the input, and attempts to predict the correct alpha. The goal of $D$ is to distinguish between real 4-channel images and fake composite images.
Another approach proposed by \cite{ren2021generative} is to optimize matting and harmonization simultaneously, implying that both techniques improve each other. 


\begin{figure*}
  \centering
  \includegraphics[scale=0.5]{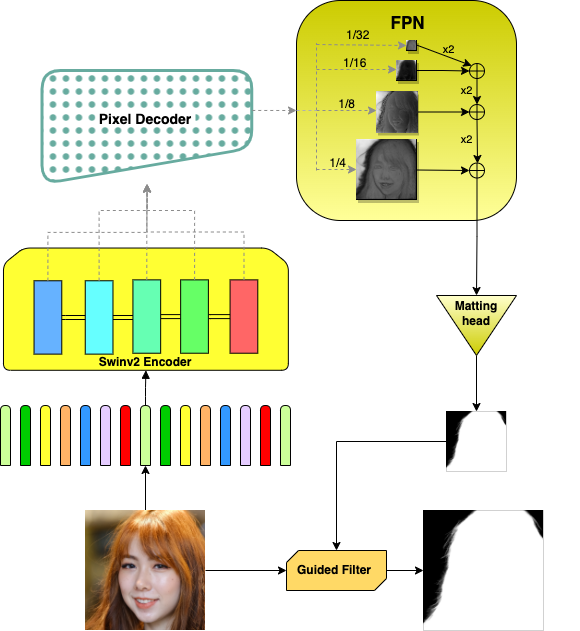}
  \caption{StyleMatte architecture}
  \label{fig: stylematte}
\end{figure*}
Labels4free~\cite{labels4free} proposes an unsupervised segmentation framework for StyleGAN-generated objects. They extract and upsample low-level features from StyleGAN2, which contains information for segmentation. They modify the generator architecture with a segmentation branch and split the generator into foreground and background networks, which leads to soft segmentation masks gaining for the foreground object in an unsupervised fashion.


FurryGAN~\cite{bae2022furrygan}, which is closest to StyleMatteGAN, introduces StyleGAN2 utilization to generate a matting mask with a focus on the hair and whiskers. They use foreground and background generators in the same manner as Labels4free~\cite{labels4free} and add a special mask predictor.
This predictor consists of two networks, coarse-grain and fine-grain, and receives input from the foreground generator. The mask generation is guided by an auxiliary mask predictor. They applied five objective constraints to the generators and produced a matting mask on $256 \times 256$.

\section{Methods}

\newcommand{\x}{\mathbf{x}}
\renewcommand{\u}{\mathbf{u}}
\renewcommand{\v}{\mathbf{v}}
\newcommand{\y}{\mathbf{y}}
\newcommand{\w}{\mathbf{w}}
\newcommand{\m}{\mathbf{m}}
\renewcommand{\a}{\mathbf{a}}
\newcommand{\bw}{\text{bw}}
\newcommand{\fw}{\text{fw}}
\newcommand{\gt}{\text{gt}}

\begin{figure*}
  \centering
  \includegraphics[width=\linewidth]{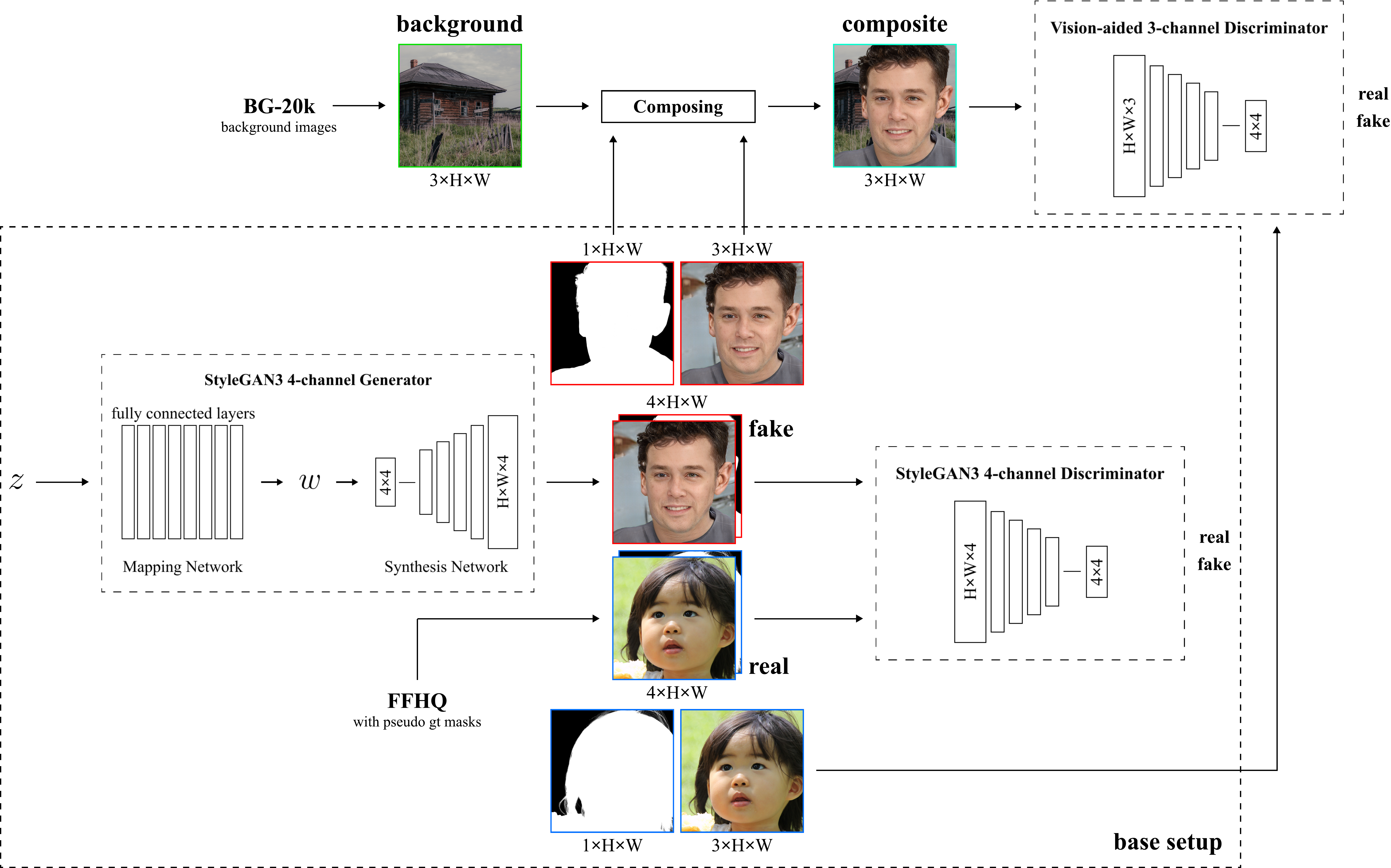}
  \caption{StyleMatteGAN architecture. Given a latent vector $z$, the RGB image and alpha mask are produced by a 4-channel generator. The obtained result and image from FFHQ with pseudo ground truth alpha mask are fed to the discriminator as fake and real, respectively. This part of the framework is our baseline method. Further, to constrain the generated 4-channel tensor, we create a composite image from a fake RGB image, mask, and random background. Then, a second Vision-aided discriminator is trained to separate the composite image from the real 3-channel image.}
  \label{fig: full}
\end{figure*}


We have two networks: StyleMatte for alpha matte pooling from images and StyleMatteGAN for image-matte pair generation. Both networks are pretrained; StyleMatte is pretrained on AM-2k and P3M-10k datasets, and StyleMatteGAN used pretrained StyleGAN-3 image generators.

StyleMatte trains on RGB images and produces alpha masks. These images are concatenated with masks and fed to StyleMatteGAN as training samples. The produced synthetic image-matte pairs are used as additional sources for matte extracting network training. As can be seen, this cycle refinement trick can be applied many times, while there is a significant improvement in the produced alpha matte (Figure~\ref{fig: improve}).

We use our networks in two domains: human portraits and animal faces. We utilize human portraits from the FFHQ~\cite{karras2019style} dataset and animal faces from AFHQv2~\cite{afhq} as a ground truth for adversarial training.

\subsection{Matting network}
\label{section:matting}

We present StyleMatte, a modern matting network design. We expect that progress in the segmentation task could lead to good matting results. We were inspired by the Mask2Former~\cite{cheng2021mask2former} design for our architecture (Figure~\ref{fig: stylematte}), and used the Swinv2~\cite{swinv2} encoder and pixel decoder. We feed four outputs from the decoder to the feature pyramid aggregation layer. To minimize the number of parameters, we upscale the smaller decoder layer and concatenate it with the next layer.

We use L1 and Laplacian losses on masks and MSE loss on composite images as objectives. For Laplacian loss, we obtain Laplacian pyramids on pairs of images and compute the sum of the absolute differences at all stages. The predicted composite image is constructed from the predicted alpha mask, foreground and background  \eqref{eq:alpha}.

We experiment with several guided filters to upscale the mask to the original size. 
This network obtains competitive results on P3M-10k and AM-2k.

\subsection{Generating image and mask}
Our goal is to generate an image and a mask together. We use RGBA image representation format. We adjust the architecture of StyleGAN3 so that it generates 4-channel images (Figure~\ref{fig: full}). We obtain matting masks for ground truth images~\cite{karras2019style, afhq} using StyleMatte. We modify the original generator's last block to produce alpha masks and the original discriminator by changing the number of input channels. The data processing functions were adjusted to these shapes accordingly.

We create a combination of different channel discriminators. Considering $G$ as a generative network,  $D$ as discriminator, we solve minimax GAN problem~\cite{gan}
\begin{equation} \label{eq:gan}
\begin{aligned}
\arg \min_{\theta_G}\max_{\theta_D} \mathbb{E} _{r \sim {p_{data}}} [logD(r)] \\ + \mathbb{E} _{z \sim \mathcal{N}} [log(1
- D(G(z))] 
\end{aligned}
\end{equation}

In case of 2 discriminators ($D_3$, $D_4$) our optimization task is modified into
\begin{equation} \label{eq:gan2d}
\begin{aligned}
\arg \min_{\theta_G}\max_{\theta_{D_3}, \theta_{D_4}} \mathbb{E} _{r \sim {p_{data}}} [log{D_3(r)} + \lambda * log{D_4(r)}] \\ + \mathbb{E} _{z \sim \mathcal{N}} [log(1
- D_3(G(z))) * (1
- D_4(G(z)))] 
\end{aligned}
\end{equation}

As in Ensembling Off-the-shelf Models for GAN Training~\cite{kumari2021ensembling}, we use a pair of discriminators. This helped converge the accurate mask generation.
We set up a generator $G$ to produce 4-channel images, three of which are color representations, and the fourth is the matte of the same object. $D_4$ attempts to distinguish RGBA outputs of $G$ from pairs of ground truth image and mask, produced by StyleMatte. The second 3-channel discriminator $D_3$ aims to maintain the perceptual stability of the generated images by tuning the mixed activation layers of the pretrained visual recognition networks. For the portraits, we set $D_3 = \sum_i Linear(U_i; \theta_i)$, where $U_i$ is $i$-th frozen activation layer of the facial parsing network, and $Linear$ is a trainable projection of this layer. We train StyleMatteGAN unsupervisedly by using only a combination of discriminators.


\section{Experiments}
The matting network is trained at 800px resolution. StyleMatteGAN experiments with portrait matting were conducted on $1024\times1024$-resolution images. Additional studies on animal faces were performed on $512\times512$, the original resolution of the AFHQ~\cite{afhq}. It is vital that we use high-resolution human portraits because it leads to accurate mask generation with better boundaries.





\begin{table*}[ht]
  \tiny
  \centering  
    \begin{tabular}{|cc|c|c|c|c|c|c|c|c|c|c|c|}
    \hline
    \multicolumn{2}{|c|}{Method}                                & LF           & HATT    & SHM             & MODNet      & GFM       & DIM*   & P3M(R)    &  P3M(S)+P3M(V) & P3M(V)  & \textbf{StyleMatte}  & \textbf{StyleMatte (+synthetic)} \\ \hline
    \multicolumn{2}{|c|}{Backbone}                              & DenseNet-201 & ResNeXt & PSPNet-50+VGG16 & MobileNetV2 & ResNet-34 & VGG16  & ResNet-34 & Swin-T            & ViTAE-S & Swin-T & Swin-T \\ \hline
    \multicolumn{1}{|c|}{\multirow{10}{*}
    {\vtop{\hbox{\strut P3M-500-P }\hbox{\strut (face-blurred)}}}}
      & SAD    & 42.95        & 25.99   & 21.56           & 13.31       & 13.20     & -      & 8.73      & 7.43              & 6.24    &     7.57 & 6.97 \\ \cline{2-13} 
    \multicolumn{1}{|c|}{}                             & MSE    & 0.0191       & 0.0054  & 0.0100          & 0.0038      & 0.0050    & -      & 0.0026    & 0.0022            & 0.0015  & 0.0024 &    0.0019  \\ \cline{2-13} 
    \multicolumn{1}{|c|}{}                             & MAD    & 0.025        & 0.015   & 0.013           & 0.008       & 0.008     & -      & 0.005     & 0.004             & 0.004   &     0.0044 &  0.004 \\ \cline{2-13} 
    \multicolumn{1}{|c|}{}                             & SAD-T  & 12.43        & 11.03   & 9.14            & 8.11        & 8.84      & 4.89   & 6.89      & 5.80              & 5.65    &  5.933 & 5.85  \\ \cline{2-13} 
    \multicolumn{1}{|c|}{}                             & MSE-T  & 0.0421       & 0.0377  & 0.0255          & 0.0258      & 0.0269    & 0.0115 & 0.0193    & 0.0151            & 0.0142  &  0.0171 & 0.0151 \\ \cline{2-13} 
    \multicolumn{1}{|c|}{}                             & MAD-T  & 0.0824       & 0.0752  & 0.0545          & 0.0563      & 0.0616    & 0.0342 & 0.0478    & 0.0395            & 0.0385  &   0.0403 & 0.0399  \\ \cline{2-13} 
    \multicolumn{1}{|c|}{}                             & SAD-FG & 18.922       & 2.575   & 0.486           & 2.777       & 0.872     & -      & 0.673     & 0.563             & 0.184   &  0.398 & 0.293  \\ \cline{2-13} 
    \multicolumn{1}{|c|}{}                             & SAD-BG & 11.595       & 12.385  & 13.098          & 2.424       & 3.487     & -      & 1.166     & 1.075             & 0.409   &  1.243  & 0.827  \\ \cline{2-13} 
    \multicolumn{1}{|c|}{}                             & Grad   & 42.19        & 14.91   & 21.24           & 16.50       & 12.58     & 4.48   & 8.22      & 11.59             & 10.94   & 13.64 & 12.37  \\ \cline{2-13} 
    \multicolumn{1}{|c|}{}                             & Conn   & 18.80        & 25.29   & 17.53           & 10.88       & 17.75     & 9.68   & 13.68     & 7.03              & 5.86    &  7.49 & 6.61       \\ \hline
    \multicolumn{1}{|c|}{\multirow{10}{*}{P3M-500-NP}} & SAD    & 32.59        & 30.53   & 20.77           & 16.70       & 15.50     & -      & 11.23     & 7.94              & 7.59   & 8.65 & 8.16 \\ \cline{2-13} 
    \multicolumn{1}{|c|}{}                             & MSE    & 0.0131       & 0.0072  & 0.0093          & 0.0051      & 0.0056    & -      & 0.0035    & 0.0021            & 0.0019 & 0.0025 & 0.0021         \\ \cline{2-13} 
    \multicolumn{1}{|c|}{}                             & MAD    & 0.0188       & 0.0176  & 0.0122          & 0.0097      & 0.0091    & -      & 0.0065    & 0.0047            & 0.0044  & 0.0049 & 0.0046    \\ \cline{2-13} 
    \multicolumn{1}{|c|}{}                             & SAD-T  & 14.53        & 13.48   & 9.14            & 9.13        & 10.16     & 5.32   & 7.65      & 6.51              & 6.60    & 6.639 &      6.753   \\ \cline{2-13} 
    \multicolumn{1}{|c|}{}                             & MSE-T  & 0.0420       & 0.0403  & 0.0255          & 0.0237      & 0.0268    & 0.0094 & 0.0173    & 0.0138            & 0.0142  &   0.0154 & 0.0146     \\ \cline{2-13} 
    \multicolumn{1}{|c|}{}                             & MAD-T  & 0.0825       & 0.0803  & 0.0545          & 0.0549      & 0.0620    & 0.0324 & 0.0466    & 0.0390            & 0.0396  &  0.0396 & 0.0395       \\ \cline{2-13} 
    \multicolumn{1}{|c|}{}                             & SAD-FG & 8.924        & 2.930   & 0.935           & 3.143       & 2.172     & -      & 1.414     & 0.439             & 0.148   &  0.617 & 0.53     \\ \cline{2-13} 
    \multicolumn{1}{|c|}{}                             & SAD-BG & 9.136        & 14.121  & 10.701          & 4.434       & 3.161     & -      & 2.165     & 0.988             & 0.838   &   1.295 & 0.881     \\ \cline{2-13} 
    \multicolumn{1}{|c|}{}                             & Grad   & 31.93        & 19.88   & 20.30           & 15.29       & 14.82     & 4.70   & 10.35     & 10.33             & 9.84    &   11.69 & 10.64     \\ \cline{2-13} 
    \multicolumn{1}{|c|}{}                             & Conn   & 19.50        & 27.42   & 17.09           & 13.81       & 18.03     & 7.70   & 12.51     & 7.25              & 6.96    &  8.215 & 7.69       \\ \hline
    \end{tabular}
    \captionsetup{justification=centering}
    \caption{Matte pulling metrics comparison ~\cite{rethink}. StyleMatte denotes our model.
    StyleMatte (+synthetic) additionally trained on synthetic dataset.}
    \label{tabl:metrics}
    
\end{table*}

\begin{table*}[ht]
    \centering
    \begin{tabular}{|c|c|c|c|c|c|c|c|c|c|}
    \hline
    \multicolumn{1}{|c|}{Method}  & GFM(r')           & GFM(r2b)    & GFM(d)             & GFM(r)      & SHM       & HATT   & LF    &  SHMC  & \textbf{StyleMatte}  
    \\  \hline
    \multicolumn{1}{|c|}{SAD}
       & 61.5 & 36.12 & 28.01 & 17.81 & 10.89 & 10.26      & 10.24 & 9.66              &    9.602  \\  
    \multicolumn{1}{|c|}{MSE}                & 0.027       & 0.0116  & 0.0055          & 0.0068      & 0.0029    & 0.0029 & 0.0028 & 0.0024            & 0.0024    \\  
    \multicolumn{1}{|c|}{MAD}                              & 0.0356        & 0.021   & 0.0161           & 0.0102       & 0.0064     & 0.0059      & 0.006     & 0.006             & 0.0055        \\ \cline{1-10} 
    \end{tabular}
    \caption{Matte pulling metrics comparison on AM-2k.}
\label{tabl:metrics_am2k}
\end{table*}

\begin{table*}[ht]
    \centering
    \begin{tabular}{|c|c|c|c|c|c|}
    \hline
    \multicolumn{1}{|c|}{Method}  & PSeg~\cite{bielski2019emergence}           & Labels4Free~\cite{labels4free} & FurryGAN~\cite{bae2022furrygan} & \textbf{StyleMatteGAN}  
    & \textbf{StyleMatteGAN (Vision-aided)} 
    \\  \hline
    \multicolumn{1}{|c|}{FFHQ}
       & 62.44 & 6.51 & 8.72 & 8.21 & 5.13  \\  
    \multicolumn{1}{|c|}{AFHQ v2}                & 12.71 & 5.19  & 6.34          & 11.31      & 4.23      \\  
    \cline{1-6} 
    \end{tabular}
    \caption{FID metrics comparison of different GAN architectures on FFHQ and AFHQ v2.}
    \label{tabl:fid}
\end{table*}

\subsection{Dataset preparation}

We utilize Fast Multi-Level Foreground Estimation~\cite{bg_extraction} to extract foregrounds. These foregrounds are blended \eqref{eq:alpha} with new backgrounds, and the resulting compositions are used as inputs for StyleMatte.
Additional background images were obtained from the BG-20k dataset~\cite{li2022matting}.

We take the original human face FFHQ dataset~\cite{karras2019style} and create a set of image masks based on our matting network.
First, we filter the dataset to contain images with only one person using an instance segmentation neural network~\cite{retinaface}. The filtration step is obligatory because some visual artifacts arise when skipping it. We obtain the results of StyleMatte for the remaining $90\%$ images as additional $\alpha$-channels. We perform an additional filtration step to get rid of some masks that are visually inconsistent with the contours of the portrait.
To do so, we extract segmentation masks $s$ with a basic semantic segmentation network~\cite{semsegm2015} pretrained on ResNet50~\cite{resnet50} and choose only those that are aligned with the matte. In other words, we leave images that fulfill the alignment agreement condition
\begin{equation} \label{eq:filter}
    \left\|(\alpha | {\alpha>\varepsilon})- s \right\|< t
\end{equation}
where $(\alpha | {\alpha>\varepsilon})$ denotes the matte mask binarized above the fixed value $\varepsilon$ and $t$ is the threshold.
We set ${\varepsilon=0.1}$ and $t=0.1$ considering $\alpha \in [0, 1]$.
Finally, we combine the resulting images and masks in the png files with $\alpha$-channels, encoded as RGBA.

To check the robustness of our method, we performed a procedure similar to of that animal faces AFHQ v2~\cite{afhq}. We use all animal categories in the COCO segmentation output, consistent with the align agreement condition~\eqref{eq:filter}.



\subsection{Training matting network}

The training experiment consists of two phases. First, we freeze the encoder parameters and pretrain the decoder at low resolution with a single L1 loss for 10 epochs. 
Next, we train StyleMatte on 800px crops utilizing the AdamW optimizer with a learning rate of 1e-5 and a weighted sum of L1, Laplacian, and composition losses. We set the weight of the composition squared error to 10, compared to 1 for linear losses. We find out that a learning rate greater than 1e-5 leads to unstable training. We use the cosine learning rate warm-up schedule and complex augmentations consisting of random crops, horizontal flips, random affine transformations, blur, random changes in hue, saturation, and value. We also conduct a background replacement with a random image. 
StyleMatte is trained on 4 Nvidia Tesla V100 GPUs
with a batch size of 16, for 100 epochs.
Finally, we tune our network to full-scale images with a batch size of 1 for one epoch.

We train our matting network on composite images.
We find that the truncation value $\psi$~\cite{stylegan3} plays a significant role in the visual distribution of our matting dataset. Although $\psi \sim 0$ generates visually pleasant images, $\psi \sim 1$ adds curly hair and face attributes. In contrast, $\psi \sim -1$ tunes images to be less hairy by adding glass. In the first stage, we train our model with a truncation value beginning from 0, increasing its absolute value in step of 0.1 and randomly changing the sign. Finally, we change it to 1 to obtain detailed images with high-frequency features.

\subsection{Synthesizing image-mask pairs}
We modify the StyleGAN3 codebase to support transparent images. The generator $G$ and discriminator $D_4$ are 4-channel modifications of StyleGAN3 versions. For the generator, we take the latest output of the synthesis layer $L14$ and adjust $toRGB$ channel weights with an additional channel. We load the existing weights from the pretrained translation-invariant config and concatenate them with the randomly initialized last layer. Surprisingly, the additional channel produced by this network generates high-definition features that contain very detailed textures, such as parts of hair and whiskers. 
Similarly, we modify the first $b.fromRGB$ channel of the pretrained discriminator so that it can input the RGBA image. Color augmentation is performed for the first 3 channels and space augmentation is applied to all four channels.
The 3-channel $D_3$ is used to classify the already blended samples from $G$, and 4-channel $D_4$ is used to classify the concatenated RGB image and alpha from $G$ and pseudo ground truth inputs from FFHQ. 

The extra discriminator ($D_3$) relies on the feature extractor from Vision-aided GAN~\cite{kumari2021ensembling}. We train a 4-channel generator with guidance from a Vision-aided discriminator and $D_4$. In our experiment, we preload weights from our base model with two discriminators and fine-tune them on the Vision-aided discriminator. We try different off-the-shelf models and chose CLIP.

StyleMatteGAN is trained on 4 NVIDIA Tesla V100 GPUs, with a batch size of 32. We set $R_1$ regularization weight $gamma = 6.6$ for human face experiments and $8.2$ for animals. 

\subsection{Evaluating results}
We use SAD (sum of absolute differences),
MSE (mean squared error), MAD (mean absolute difference) for whole image and its trimap-based parts. Also we compute gradient and connectivity errors for comparison (Table~\ref{tabl:metrics}). We conducted experiments with original datasets and datasets mixed with synthetic output. 
As far as synthetic data contains unblurred portraits, the improvement on P3M-500-P (blurred faces) was less significant than on P3M-500-NP.

We evaluate the quality of the images produced by StyleMatteGAN and the consistency of the matte masks applied to portraits.
In addition, we match the results with those of existing studies. However, this comparison has some limitations. First, the existing works do not produce high-definition masks, and FurryGAN for portrait alpha masks does not have a published network. Therefore, we use Fréchet inception distance metrics from this study ~\cite{bae2022furrygan} for comparison. The results are presented in Table~\ref{tabl:fid}. StyleMatteGAN maintains better performance on FID because of its novel architecture.
For $\alpha$ mask consistency with image $C$ we tested several matte-pulling networks. In this evaluation, we set the RGB image $C$ produced by the StyleMatteGAN as the input to these networks. We compare the output of this network with the generated $\alpha$ mask using the Mean Square Error (MSE) and Mean Absolute Difference (MAD).

We also evaluate our model on AM-2k dataset, which contains images with animals. StyleMatte obtains state-of-the-art results on them (Table~\ref{tabl:metrics_am2k}).

\subsection{Ablation Study}

\noindent \textbf{StyleMatte design.} Feature pyramid network (Figure ~\ref{fig: stylematte}) reduces SAD on P3M-500-P by 0.5, and P3M-500-NP by 0.9 compared to plain output from last decoder stage. Fast guided filter improves high-frequent features quality and SAD by 0.6 compared to bilinear upscale, however it leads to pixel artifacts.

\noindent \textbf{Default vs. Vision-aided GAN discriminator.} \cite{kumari2021ensembling} proposes to extend the discriminator ensemble with pretrained feature networks which improves our method (Table~\ref{tabl:fid}).


\section{Discussion}
\subsection{Limitations and future works}
The result of our work is strongly correlated to the domain of the training set. Most of the generated portrait images contain fixed frontal head pose,
that is why StyleMatteGAN cannot generate images from above or beyond. 
The actual development of StyleGAN-based networks reveals additional tasks our network could solve. 
The improvement by using 3D-Aware generators~\cite{3d-aware} could increase the flexibility of generated poses.

The intuition of using StyleGAN3 came from its capability to generate high-frequency features. It solved the aliasing problem of its predecessor, which brings an effect of the generated image being glued from patches. This area is widely explored, and we can use all enhancements described in related works~\cite{stylefacev, gan3d}. Recent studies show that diffusion models~\cite{diffusion} get better results in diversity and quality, however we could not find suitable pipeline for our task and their usage is time-consuming.

Obtaining of alpha mattes doesn't guarantee visual pleasant image translation. In this work we don't focus on background removal task utilizing existing solution~\cite{bg_extraction}.

\subsection{Ethical consideration}
We are aware of misbehavior, i.e., the usage of generated images referred to as ``deepfakes'' in an improper way. This case is common for all StyleGAN-based solutions. We briefly investigated the ``deepfake'' detection networks and found out that they focused mainly on portraits and could easily detect such images with an accuracy of more than $98\%$ \cite{deepfakedetection}. On the other hand, our contribution could play a significant role in the privacy-preserving task. In itself, it provides a large number of images that could not enclose any person's identity. It also can be used as a source of anonymization.

\subsection{Conclusion}

We present StyleMatte for universal image matting and StyleMatteGAN for matting dataset enrichment. StyleMatte obtains top results on P3M-10k and AM-2k datasets, outperforming many state-of-the-art models and remaining easy to adapt to different backbones. StyleMatteGAN helps expand small datasets to get enough training image-mask pairs. We analyzed the possibilities of generating pseudo ground truth matting masks in conjunction with synthetic images. We applied different approaches to keep realistic pictures while keeping the masks close to them. To the best of our knowledge, our work produced the first high-resolution portraits with matte masks. We also show that this data is sufficient to improve matting networks.


\newpage
{\small
\bibliographystyle{ieee_fullname}

\begin{thebibliography}{10}
\itemsep=-1pt

\bibitem{labels4free}
Rameen Abdal, Peihao Zhu, Niloy~J Mitra, and Peter Wonka.
\newblock Labels4free: Unsupervised segmentation using stylegan.
\newblock In {\em Proceedings of the IEEE/CVF International Conference on
  Computer Vision}, pages 13970--13979, 2021.

\bibitem{bae2022furrygan}
Jeongmin Bae, Mingi Kwon, and Youngjung Uh.
\newblock Furrygan: High quality foreground-aware image synthesis.
\newblock In {\em European Conference on Computer Vision}, pages 696--712.
  Springer, 2022.

\bibitem{bielski2019emergence}
Adam Bielski and Paolo Favaro.
\newblock Emergence of object segmentation in perturbed generative models.
\newblock {\em Advances in Neural Information Processing Systems}, 32, 2019.

\bibitem{gan3d}
Eric~R Chan, Connor~Z Lin, Matthew~A Chan, Koki Nagano, Boxiao Pan, Shalini
  De~Mello, Orazio Gallo, Leonidas~J Guibas, Jonathan Tremblay, Sameh Khamis,
  et~al.
\newblock Efficient geometry-aware 3d generative adversarial networks.
\newblock In {\em Proceedings of the IEEE/CVF Conference on Computer Vision and
  Pattern Recognition}, pages 16123--16133, 2022.

\bibitem{cheng2021mask2former}
Bowen Cheng, Ishan Misra, Alexander~G Schwing, Alexander Kirillov, and Rohit
  Girdhar.
\newblock Masked-attention mask transformer for universal image segmentation.
\newblock In {\em Proceedings of the IEEE/CVF Conference on Computer Vision and
  Pattern Recognition}, pages 1290--1299, 2022.

\bibitem{afhq}
Yunjey Choi, Youngjung Uh, Jaejun Yoo, and Jung-Woo Ha.
\newblock Stargan v2: Diverse image synthesis for multiple domains.
\newblock In {\em Proceedings of the IEEE/CVF conference on computer vision and
  pattern recognition}, pages 8188--8197, 2020.

\bibitem{retinaface}
Jiankang Deng, Jia Guo, Evangelos Ververas, Irene Kotsia, and Stefanos
  Zafeiriou.
\newblock Retinaface: Single-shot multi-level face localisation in the wild.
\newblock In {\em Proceedings of the IEEE/CVF conference on computer vision and
  pattern recognition}, pages 5203--5212, 2020.

\bibitem{dosovitskiy2020image}
Alexey Dosovitskiy, Lucas Beyer, Alexander Kolesnikov, Dirk Weissenborn,
  Xiaohua Zhai, Thomas Unterthiner, Mostafa Dehghani, Matthias Minderer, Georg
  Heigold, Sylvain Gelly, et~al.
\newblock An image is worth 16x16 words: Transformers for image recognition at
  scale.
\newblock {\em arXiv preprint arXiv:2010.11929}, 2020.

\bibitem{bg_extraction}
Thomas Germer, Tobias Uelwer, Stefan Conrad, and Stefan Harmeling.
\newblock Fast multi-level foreground estimation.
\newblock {\em CoRR}, abs/2006.14970, 2020.

\bibitem{gan}
Ian Goodfellow, Jean Pouget-Abadie, Mehdi Mirza, Bing Xu, David Warde-Farley,
  Sherjil Ozair, Aaron Courville, and Yoshua Bengio.
\newblock Generative adversarial networks.
\newblock {\em Communications of the ACM}, 63(11):139--144, 2020.

\bibitem{resnet50}
Kaiming He, Xiangyu Zhang, Shaoqing Ren, and Jian Sun.
\newblock Deep residual learning for image recognition.
\newblock {\em CoRR}, abs/1512.03385, 2015.

\bibitem{diffusion}
Jonathan Ho, Ajay Jain, and Pieter Abbeel.
\newblock Denoising diffusion probabilistic models.
\newblock {\em arXiv preprint arxiv:2006.11239}, 2020.

\bibitem{stylegan3}
Tero Karras, Miika Aittala, Samuli Laine, Erik H{\"a}rk{\"o}nen, Janne
  Hellsten, Jaakko Lehtinen, and Timo Aila.
\newblock Alias-free generative adversarial networks.
\newblock {\em Advances in Neural Information Processing Systems}, 34:852--863,
  2021.

\bibitem{karras2019style}
Tero Karras, Samuli Laine, and Timo Aila.
\newblock A style-based generator architecture for generative adversarial
  networks.
\newblock In {\em Proceedings of the IEEE/CVF conference on computer vision and
  pattern recognition}, pages 4401--4410, 2019.

\bibitem{MODNet}
Zhanghan Ke, Jiayu Sun, Kaican Li, Qiong Yan, and Rynson~W.H. Lau.
\newblock Modnet: Real-time trimap-free portrait matting via objective
  decomposition.
\newblock In {\em AAAI}, 2022.

\bibitem{kumari2021ensembling}
Nupur Kumari, Richard Zhang, Eli Shechtman, and Jun-Yan Zhu.
\newblock Ensembling off-the-shelf models for gan training.
\newblock In {\em Proceedings of the IEEE/CVF Conference on Computer Vision and
  Pattern Recognition (CVPR)}, June 2022.

\bibitem{semanticGAN}
Daiqing Li, Junlin Yang, Karsten Kreis, Antonio Torralba, and Sanja Fidler.
\newblock Semantic segmentation with generative models: Semi-supervised
  learning and strong out-of-domain generalization.
\newblock In {\em Conference on Computer Vision and Pattern Recognition
  (CVPR)}, 2021.

\bibitem{vitae}
Jizhizi Li, Sihan Ma, Jing Zhang, and Dacheng Tao.
\newblock Privacy-preserving portrait matting.
\newblock In {\em {MM} '21: {ACM} Multimedia Conference, Virtual Event, China,
  October 20 - 24, 2021}, pages 3501--3509, 2021.

\bibitem{p3m}
Jizhizi Li, Sihan Ma, Jing Zhang, and Dacheng Tao.
\newblock Privacy-preserving portrait matting.
\newblock In {\em Proceedings of the 29th ACM International Conference on
  Multimedia}, MM '21, page 3501–3509, New York, NY, USA, 2021. Association
  for Computing Machinery.

\bibitem{li2022matting}
Jizhizi Li, Jing Zhang, Stephen~J Maybank, and Dacheng Tao.
\newblock Bridging composite and real: Towards end-to-end deep image matting.
\newblock {\em International Journal of Computer Vision}, 2022.

\bibitem{rvm}
Shanchuan Lin, Linjie Yang, Imran Saleemi, and Soumyadip Sengupta.
\newblock Robust high-resolution video matting with temporal guidance, 2021.

\bibitem{swinv2}
Ze Liu, Han Hu, Yutong Lin, Zhuliang Yao, Zhenda Xie, Yixuan Wei, Jia Ning, Yue
  Cao, Zheng Zhang, Li Dong, Furu Wei, and Baining Guo.
\newblock Swin transformer {V2:} scaling up capacity and resolution.
\newblock {\em CoRR}, abs/2111.09883, 2021.

\bibitem{semsegm2015}
Jonathan Long, Evan Shelhamer, and Trevor Darrell.
\newblock Fully convolutional networks for semantic segmentation.
\newblock In {\em Proceedings of the IEEE conference on computer vision and
  pattern recognition}, pages 3431--3440, 2015.

\bibitem{lutz2018alphagan}
Sebastian Lutz, Konstantinos Amplianitis, and Aljosa Smolic.
\newblock Alphagan: Generative adversarial networks for natural image matting.
\newblock {\em arXiv preprint arXiv:1807.10088}, 2018.

\bibitem{rethink}
Sihan Ma, Jizhizi Li, Jing Zhang, He Zhang, and Dacheng Tao.
\newblock Rethinking portrait matting with privacy preserving, 2023.

\bibitem{Park_2022_CVPR}
GyuTae Park, SungJoon Son, JaeYoung Yoo, SeHo Kim, and Nojun Kwak.
\newblock Matteformer: Transformer-based image matting via prior-tokens.
\newblock In {\em Proceedings of the IEEE/CVF Conference on Computer Vision and
  Pattern Recognition (CVPR)}, pages 11696--11706, June 2022.

\bibitem{stylefacev}
Haonan Qiu, Yuming Jiang, Hang Zhou, Wayne Wu, and Ziwei Liu.
\newblock Stylefacev: Face video generation via decomposing and recomposing
  pretrained stylegan3.
\newblock {\em arXiv preprint arXiv:2208.07862}, 2022.

\bibitem{ren2021generative}
Xuqian Ren, Yifan Liu, and Chunlei Song.
\newblock A generative adversarial framework for optimizing image matting and
  harmonization simultaneously.
\newblock In {\em 2021 IEEE International Conference on Image Processing
  (ICIP)}, pages 1354--1358. IEEE, 2021.

\bibitem{sengupta2020background}
Soumyadip Sengupta, Vivek Jayaram, Brian Curless, Steven~M Seitz, and Ira
  Kemelmacher-Shlizerman.
\newblock Background matting: The world is your green screen.
\newblock In {\em Proceedings of the IEEE/CVF Conference on Computer Vision and
  Pattern Recognition}, pages 2291--2300, 2020.

\bibitem{deepportrait}
Xiaoyong Shen, Xin Tao, Hongyun Gao, Chao Zhou, and Jiaya Jia.
\newblock Deep automatic portrait matting.
\newblock In {\em European conference on computer vision}, pages 92--107.
  Springer, 2016.

\bibitem{smith1996blue}
Alvy~Ray Smith and James~F Blinn.
\newblock Blue screen matting.
\newblock In {\em Proceedings of the 23rd annual conference on Computer
  graphics and interactive techniques}, pages 259--268, 1996.

\bibitem{segformer}
Enze Xie, Wenhai Wang, Zhiding Yu, Anima Anandkumar, Jose~M Alvarez, and Ping
  Luo.
\newblock Segformer: Simple and efficient design for semantic segmentation with
  transformers.
\newblock {\em Advances in Neural Information Processing Systems},
  34:12077--12090, 2021.

\bibitem{alphamatte}
D. Yaman, H. Ekenel, and A. Waibel.
\newblock Alpha matte generation from single input for portrait matting.
\newblock In {\em 2022 IEEE/CVF Conference on Computer Vision and Pattern
  Recognition Workshops (CVPRW)}, pages 695--704, 2022.

\bibitem{mgm}
Qihang Yu, Jianming Zhang, He Zhang, Yilin Wang, Zhe Lin, Ning Xu, Yutong Bai,
  and Alan Yuille.
\newblock Mask guided matting via progressive refinement network.
\newblock {\em arXiv preprint arXiv:2012.06722}, 2020.

\bibitem{deepfakedetection}
Hanqing Zhao, Wenbo Zhou, Dongdong Chen, Tianyi Wei, Weiming Zhang, and Nenghai
  Yu.
\newblock Multi-attentional deepfake detection.
\newblock In {\em Proceedings of the IEEE/CVF conference on computer vision and
  pattern recognition}, pages 2185--2194, 2021.

\bibitem{3d-aware}
Xiaoming Zhao, Fangchang Ma, David Güera, Zhile Ren, Alexander~G. Schwing, and
  Alex Colburn.
\newblock Generative multiplane images: Making a 2d gan 3d-aware.
\newblock In {\em Proc. ECCV}, 2022.

\end{thebibliography}

}
\end{document}